\documentclass[letterpaper]{article} 
\usepackage{aaai2026}  
\usepackage{times}  
\usepackage{helvet}  
\usepackage{courier}  
\usepackage[hyphens]{url}  
\usepackage{graphicx} 
\usepackage{natbib}  
\usepackage{caption} 

\usepackage{colortbl}
\usepackage{xcolor}
\usepackage{booktabs}
\usepackage{multirow}
\usepackage{adjustbox}
\usepackage{makecell}
\usepackage{array}
\usepackage{amsmath, bm}
\usepackage{arydshln}
\usepackage{amssymb}
\usepackage{tabularx}
\usepackage{hyperref}
\usepackage{graphicx}
\definecolor{lightblue}{rgb}{0.9,0.95,1}

\usepackage{pifont}
\usepackage{xcolor}
\newcommand{\xmark}{\textcolor{red}{\ding{55}}}
\newcommand{\bmark}{\textcolor{blue}{\ding{51}}}
\usepackage{algorithm}
\usepackage{algorithmic}

\usepackage{algorithm}
\usepackage{algorithmic}

\usepackage{newfloat}
\usepackage{listings}
\DeclareCaptionStyle{ruled}{labelfont=normalfont,labelsep=colon,strut=off} 
\floatstyle{ruled}
\newfloat{listing}{tb}{lst}{}
\floatname{listing}{Listing}
\title{MoCHA: Advanced Vision-Language Reasoning with MoE Connector and Hierarchical Group Attention}
\author{
    Yuqi Pang$^{1,2}$,
    Bowen Yang$^{1,2}$,
    Yun Cao$^{1,2}$\equalcontrib,
    Rong Fan$^{1,2}$,
    Xiaoyu Li$^{1,2}$,
    Chen He$^{1,2}$
}
\affiliations{
    \textsuperscript{\rm 1}Institute of Information Engineering, Chinese Academy of Sciences, Beijing, China\\
    \textsuperscript{\rm 2}School of Cyber Security, University of Chinese Academy of Sciences, Beijing, China\\

\{pangyuqi, yangbowen, caoyun, fanrong, lixiaoyu, hechen\}@iie.ac.cn
}

\usepackage{bibentry}

\begin{document}

\maketitle

\begin{abstract}
Vision large language models (VLLMs) are focusing primarily on handling complex and fine-grained visual information by incorporating advanced vision encoders and scaling up visual models. However, these approaches face high training and inference costs, as well as challenges in extracting visual details, effectively bridging across modalities. In this work, we propose a novel visual framework, \textbf{\textit{MoCHA}}, to address these issues. Our framework integrates four vision backbones (i.e., CLIP, SigLIP, DINOv2 and ConvNeXt) to extract complementary visual features and is equipped with a sparse \textbf{\textit{M}}ixture \textbf{\textit{o}}f Experts \textbf{\textit{C}}onnectors (MoECs) module to dynamically select experts tailored to different visual dimensions. To mitigate redundant or insufficient use of the visual information encoded by the MoECs module, we further design a \textbf{\textit{H}}ierarchical Group \textbf{\textit{A}}ttention (HGA) with intra- and inter-group operations and an adaptive gating strategy for encoded visual features. We train MoCHA on two mainstream LLMs (e.g., Phi2-2.7B and Vicuna-7B) and evaluate their performance across various benchmarks. Notably, MoCHA outperforms state-of-the-art open-weight models on various tasks. For example, compared to CuMo (Mistral-7B), our MoCHA (Phi2-2.7B) presents outstanding abilities to mitigate hallucination by showing improvements of 3.25\% in POPE and to follow visual instructions by raising 153 points on MME. Finally, ablation studies further confirm the effectiveness and robustness of the proposed MoECs and HGA in improving the overall performance of MoCHA. The code is available at \url{https://github.com/Pbhgit/MoCHA}.

\end{abstract}

\section{Introduction}
Vision Large Language Models (VLLMs)~\cite{liu2024llavanext,wang2024qwen2vlenhancingvisionlanguagemodels, bai2025qwen25vltechnicalreport} have shown promising results in multimodal reasoning tasks by integrating visual encoders with Large Language Models (LLM)~\cite{gpt,openai2024gpt4o,touvron2023llama}.
However, the large model size and extensive training data present significant computational challenges. Additionally, excessive early reliance on language in VLLMs may act as a shortcut~\cite{geirhos2020shortcut,yuksekgonul2023visionlanguagemodelsbehavelike} that results in suboptimal learning of effective visual representations. 
Therefore, exploring a VLLM that balances efficient visual processing and performance is a crucial topic.

Previous works have focused on strengthening vision encoders by utilizing extensive and higher-quality visual instruction data~\cite{zheng2023judgingllmasajudgemtbenchchatbot}, scaling up model size~\cite{chen2024internvl}, or decomposing images into low-resolution patches~\cite{shi2024needlargervisionmodels}.
While effective, these approaches are inherently limited by high computational costs and the loss of important details, such as small objects, which in turn can result in hallucinations.
Furthermore, Cambrian-1~\cite{tong2024cambrian1fullyopenvisioncentric}, designed with a vision centric approach, introduces a novel paradigm for vision-language fusion through its combination of multiple vision encoders and spatial vision aggregator (SVA), but faces difficulties due to the lack of dynamic feature fusion and shortcomings in fine-grained perception. 
Specifically, visual features are inherently high-dimensional and uncertain, encompassing objects, scenes, attributes, and spatial relationships. A single vision encoder, or a limited set thereof, is insufficient to comprehensively and adaptively capture these diverse aspects.
Sparse Mixture of Experts (MoE)~\cite{fedus2022switch} has emerged as an efficient solution that dynamically activates only a subset of expert networks for each input to improve scalability and specialization.
Although MoE has been widely adopted in state-of-the-art open-source LLMs, its potential remains largely underexplored in the design of connectors and the broader architecture of VLLMs.

To address the above limitations of existing vision encoders and enhance fine-grained visual understanding, we propose \textbf{\textit{MoCHA}}, a novel framework integrates multiple vision backbones under a sparse \textbf{\textit{M}}ixture \textbf{\textit{o}}f Experts \textbf{\textit{C}}onnector (MoECs) module and a \textbf{\textit{H}}ierarchical Group \textbf{\textit{A}}ttention (HGA) component for vision-extensive reasoning tasks.
Specifically, MoCHA consists of multiple vision encoders connected by a mixture of experts module and hierarchical group attention, as illustrated in Figure~\ref{fig:arch}. 
First, we leverage four distinct yet complementary vision backbones, including CLIP~\cite{radford2021learningtransferablevisualmodels}, SigLIP~\cite{zhai2023sigmoidlosslanguageimage}, DINOv2~\cite{oquab2024dinov2learningrobustvisual}, and ConvNeXt~\cite{liu2022convnet2020s}, to extract diverse and robust image features. However, the efficient integration of these heterogeneous visual signals into a unified vision-language framework remains a challenge. To address this, we propose the mixture of experts connectors (MoECs) module, in which each connector
is implemented as a Top-K sparsely-gated MoE for dynamic expert selection across visual dimensions. This facilitates efficient cross-modal interaction and reduces training costs. 
Sequential concatenation of features from multiple vision encoders along the token dimension often leads to redundancy and feature overlap. We further design hierarchical group attention (HGA), which fuses features through intra- and inter-group attention. By adaptive gating mechanism, we dynamically balance the aggregated and original features to produce the image representation without additional parameters.
Our MoCHA is fully trained on open-source datasets and demonstrates superior performance compared to other VLLMs of similar size. In detail, we implement MoCHA with a two-stage training process on Phi2-2.7B and Vicuna-7B-v1.5 LLMs, achieving higher scores than existing open-weight VLLMs on mainstream vision-language benchmarks. Notably, our 3B-sized MoCHA shows a 3.25\% reduction in hallucination on POPE and delivers a 153-point improvement on general visual tasks on MME, outperforming the larger CuMo-7B model~\cite{li2024cumoscalingmultimodalllm}. This demonstrates the efficacy of our MoCHA in mitigating
VLM hallucinations and enhancing general perceptual abilities. We further conduct ablation studies to investigate the contribution of each component to overall performance.

\section{Related Work}
\subsection{Large Pre-trained Vision Models}
The advent of pre-trained Vision Transformers (ViT)~\cite{dosovitskiy2021imageworth16x16words} has significantly propelled the advancement of computer vision. As the original CLIP adopted by conventional visual instruction tuning approaches is trained on noisy image-text pairs, it exhibits specific visual shortcomings, and thus stronger backbones have been introduced to LLMs. For example, SigLIP introduced pairwise sigmoid loss during training, enabling the vision encoder to demonstrate more advanced visual perception capabilities. Some concurrent works have leveraged external assistance from vision-only, self-supervised models such as DINOv2, MoCo-v3~\cite{he2021maskedautoencodersscalablevision}, and other advanced frameworks ~\cite{caron2021emergingpropertiesselfsupervisedvision,chen2024internvl}, as well as domain-specific expert models. Additionally, ConvNeXt contributes to this progress by incorporating a purely convolutional backbone that achieves Transformer-level performance with improved computational efficiency. Collectively, these approaches foster robust multimodal integration and significantly enhance visual reasoning capabilities in contemporary VLLM frameworks by leveraging complementary vision encoders to enrich visual representations and improve fine-grained perception.
\subsection{Mixture of Experts}
Mixture of Experts (MoE) proposes a group of expert networks to handle specific tasks, with a gating network selecting the appropriate experts. Subsequent research on language MoE models has further expanded large MoE-based language models, improving expert stability and load balancing. ST-MoE~\cite{zoph2022stmoedesigningstabletransferable} employs load-balancing loss and router-z loss to ensure an even distribution of experts. The recently popular DeepSeek-V3 adopts the DeepSeekMoE~\cite{dai2024deepseekmoeultimateexpertspecialization} architecture, leveraging shared experts to capture common knowledge and reduce redundancy in routed experts. MoE’s success has also extended to the vision domain. LIMoE~\cite{mustafa2022multimodalcontrastivelearninglimoe} replaces dense MLP layers with MoE layers in CLIP, improving zero-shot image classification. AdaMV-MoE~\cite{liu2024adamolefinetuninglargelanguage} introduces an adaptive MoE framework for multi-task learning. Furthermore, CuMo~\cite{li2024cumoscalingmultimodalllm} integrates Co-upcycled Top-K sparse-gating MoE blocks into the visual encoder, MLP connector, and language model, enhancing VLLM inference performance.

\section{Methodology}
\subsection{Overview}
\begin{figure*}[!htbp]
  \centering
  \includegraphics[width=0.89\linewidth]{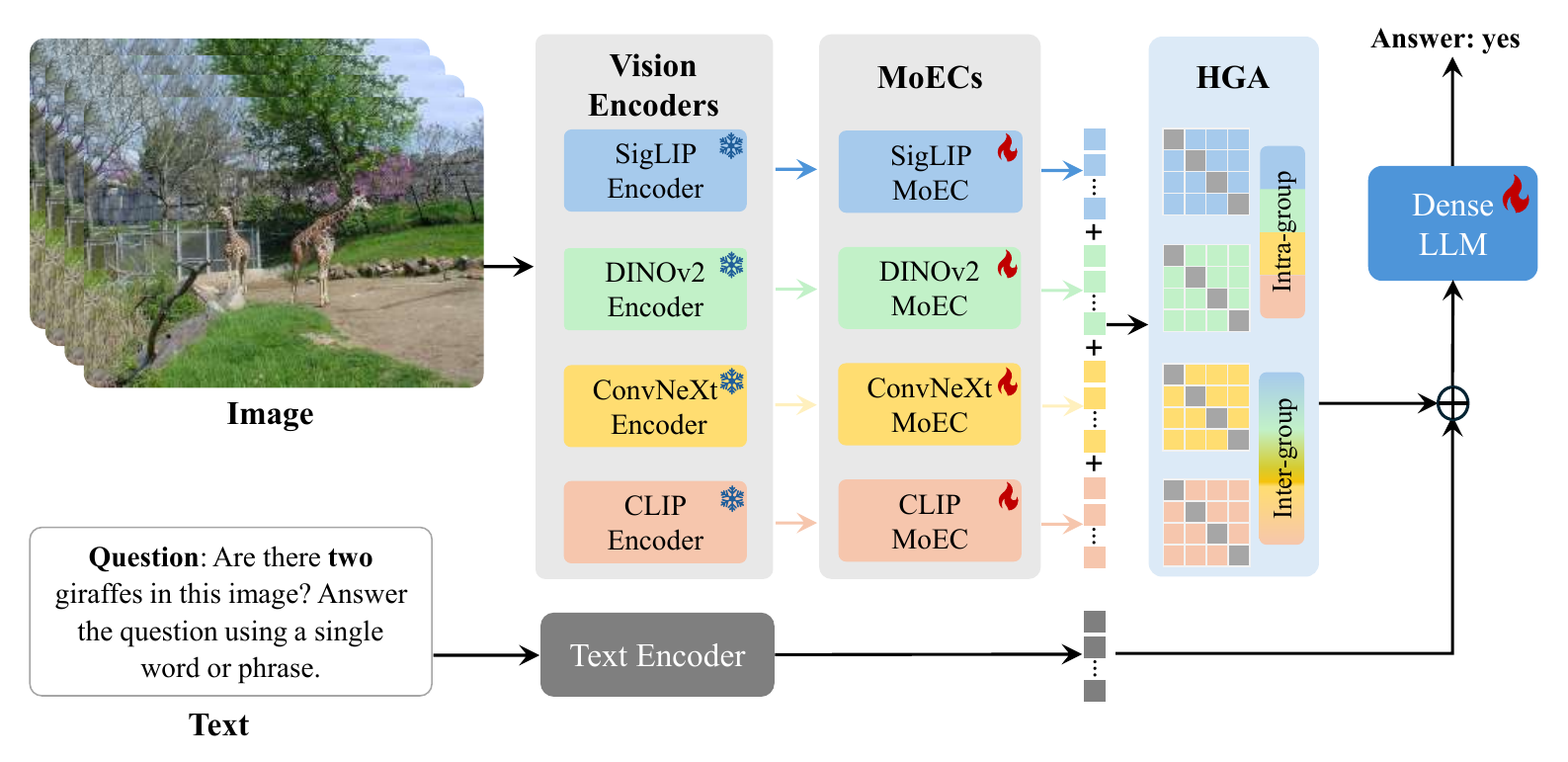}
  \caption{Architecture of MoCHA. MoCHA integrates multiple vision encoders (i.e., CLIP, SigLIP, ConvNeXt, and DINOv2), and consists of a Mixture of Experts Connectors (MoECs) module and Hierarchical Group Attention (HGA) module.
  }
       \label{fig:arch}
\end{figure*}

To address a critical gap where current VLLMs still struggle to solve vision-extensive multimodal tasks, we present our framework, MoCHA, in this section.
Building upon the mainstream LLaVA architecture, MoCHA moves beyond the original CLIP encoder by integrating a diverse set of vision encoders, spanning different architectures and pre-training objectives. This enables the model to effectively capture and process visual information across varying levels of granularity.
As illustrated in Figure~\ref{fig:arch}, our framework consists of multiple vision encoders (i.e., OpenAI CLIP ViT-L/14@336, SigLIP ViT-L/16@384, OpenCLIP ConvNeXt-XXL@1024 and DINOv2 ViT-L/14@336) connected by a mixture of experts module and hierarchical group attention.

\subsection{Selection of Vision Encoders}
Although language-supervised models outperform self-supervised and other models across all benchmark categories, Cambrian-1~\cite{tong2024cambrian1fullyopenvisioncentric} highlights that well-trained self-supervised models like DINOv2 are capable of achieving competitive performance in vision-centric tasks. In particular, high-resolution models excel in chart-related and other vision-centric benchmarks while also demonstrating robust performance across both general VQA and knowledge-based VQA. While ViTs remain the dominant architecture, ConvNet-based models such as OpenCLIP ConvNeXt are well-suited for high-resolution image processing~\cite{vishniakov2024convnetvstransformersupervised} delivering outstanding results on OCR, chart interpretation, and other vision-centric benchmarks. 

Each vision encoder excels in different aspects of VLLM performance. More analysis on the vision encoders can be found in the \textbf{Appendix A}.
SigLIP consistently excels in vision tasks due to its strong cross-modal semantic understanding and stability, and is therefore used as the baseline vision encoder. In this study, we explore the potential of combining CLIP, SigLIP, ConvNeXt, and DINOv2 to leverage their distinctive representations, aiming to enhance the exploitation of heterogeneous visual features. 
\subsection{Complementary Visual Feature Extraction}
To address the challenge of varying input resolutions across different vision encoders, a common approach is to employ interpolation for aligning feature dimensions.
However, this process can result in information loss. To mitigate this, we opt to forgo interpolation and instead ensure that the output features of all encoders share consistent channel dimensions. In our experiments, we apply pooling to the features produced by ConvNeXt-XXL@1024 to standardize the channel dimensions, while maintaining the configurations of the other encoders unchanged. Each vision encoder processes the image $I$ independently:
\begin{equation}
  \begin{aligned}
    \textbf{SigLIP}(I) &= e_s \in \mathbb{R}^{s\times D_s}, \\
    \textbf{DINOv2}(I) &= e_d \in \mathbb{R}^{d\times D_d}, \\
    \textbf{ConvNeXt}(I) &= e_n \in \mathbb{R}^{n\times D_n}, \\
    \textbf{CLIP}(I) &= e_c \in \mathbb{R}^{c\times D_c},
  \end{aligned}
\end{equation}
here, $e_s$, $e_d$, $e_n$ and $e_c$ represent the embeddings of image $I$ generated by the four vision encoders. $s$, $d$, $n$ and $c$ denote the number of tokens for each encoder, and $D_s$, $D_d$, $D_n$ and $D_c$ represents the dimension of the feature channel.
\begin{figure}[htbp]
  \centering
  \includegraphics[width=0.99\columnwidth]{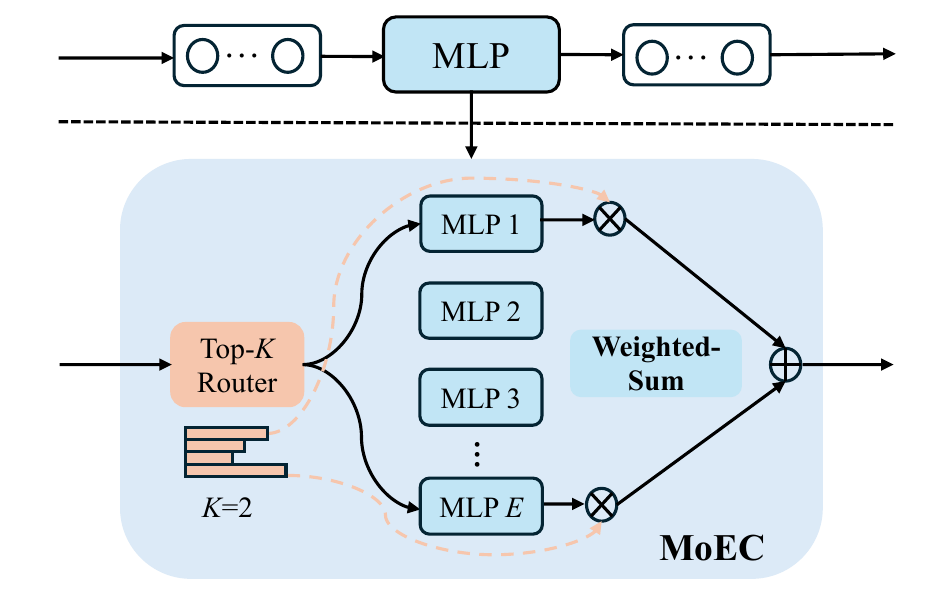}
  \caption{Sparse Top-$K$ MoEC block. Replacing MLP connector with MoEC for visual-text token alignment.}
       \label{fig:moe}
\end{figure}

\subsection{Mixture of Experts Connectors (MoECs)}
The output of LLMs is typically semantically uniform and exhibits minimal variation across dimensions, which renders specialized experts unnecessary. In contrast, vision models generate diverse information across multiple dimensions, such as objects, scenes, attributes, and spatial relationships, introducing greater uncertainty and variability. This complexity makes them well-suited for processing by multiple experts, each specializing in different visual aspects.

As illustrated in Figure~\ref{fig:moe}, we introduce an MoE into the connector of VLLM. We refer to this design as the Mixture of Experts Connector (MoEC).
MoEC preserves the stability and generality of vision-language architectures while effectively capturing the multi-dimensional and uncertain nature of visual information. 
MoEC retains the core functionality of the connector by projecting visual tokens into the word embedding space and aligning them with textual representations. 

By leveraging the connector’s small scale, flexibility and focus on intermodal fusion, MoEC dynamically selects expert strategies tailored to different visual dimensions, thereby enhancing the efficiency of modality interaction and reducing training complexity. 

Previous mainstream approaches~\cite{shazeer2017outrageouslylargeneuralnetworks} replace dense MLP blocks with sparsely-gated mixture of experts blocks. As an example, consider the SigLIP vision encoder embedding. Given input $e_s\in \mathbb{R}^{s\times D_s}$ and an MLP block, the hidden representation is computed as: 
\begin{equation}
    e_{s,h}=\text{MLP}(e_s)\in \mathbb{R}^{s\times D},
\end{equation}
To scale up the model with multiple MLP blocks in parallel, a sparse MoE block incorporates a router network that selects the Top-$K$ experts from a total of $E$  experts, as illustrated in Figure~\ref{fig:moe}. The router network uses a linear layer to compute a normalized weight matrix based on the inputs $e_s$, enabling expert selection:
\begin{equation}
  W_s=\text{Softmax}(\text{Linear}(e_s))\in \mathbb{R}^{s\times E},
\end{equation}
Based on $W_s$, the Top-$K$ experts are selected for each token, and the corresponding weights are re-normalized as:
\begin{equation}
 W_{s,K}=\text{Softmax}(\text{Top}K(W_s))\in \mathbb{R}^{s\times K},
\end{equation}
Each selected expert corresponds to an MLP block, and the final hidden representation is obtained through a re-weighted sum:
\begin{equation}
e_{s,h}=\sum_{i}^{K}W_{s,K}^{i}\circ \text{MLP}_i(e_s)\in \mathbb{R}^{s\times D}.
\end{equation}

The hidden representation of SigLIP MoEC maintains the same dimensionality as that of a single dense MLP block. Similarly, the hidden representations for DINOv2 MoEC, ConvNeXt MoEC, and CLIP MoEC are obtained in the same way:
$e_{d,h}\in \mathbb{R}^{d\times D}$, $e_{n,h}\in \mathbb{R}^{n\times D}$, $e_{c,h}\in \mathbb{R}^{c\times D}$.

\subsubsection{Sequence Append (Token Dimension): }
While channel concatenation~\cite{lin2023sphinxjointmixingweights} achieves performance comparable to sequence append~\cite{liu2024prismervisionlanguagemodelmultitask,kar2024bravebroadeningvisualencoding,fan2024mousipolyvisualexpertvisionlanguagemodels}, it requires strict spatial alignment of encoder outputs, rendering it less compatible with encoders of differing resolutions and architectures. 
Sequence append along the token dimension obviates token count alignment and removes the requirement for interpolation, resampling, or padding. As a result, encoders with different resolutions, patch sizes, or feature map dimensions can contribute arbitrary numbers of tokens. In this study, we adopt sequence append, concatenating features from different encoders directly along the token dimension:
\begin{equation}
  \begin{split}
    X_{\text{in}} = (\text{Concatenate}[e_{s,h}, e_{d,h},
    e_{n,h}, e_{c,h}], \text{ dim=token}) 
    \\\in \mathbb{R}^{N\times D},
  \end{split}
\end{equation}
where $s+d+n+c=N$.
\begin{figure}[!ht]
  \centering
  \includegraphics[width=0.89\columnwidth]{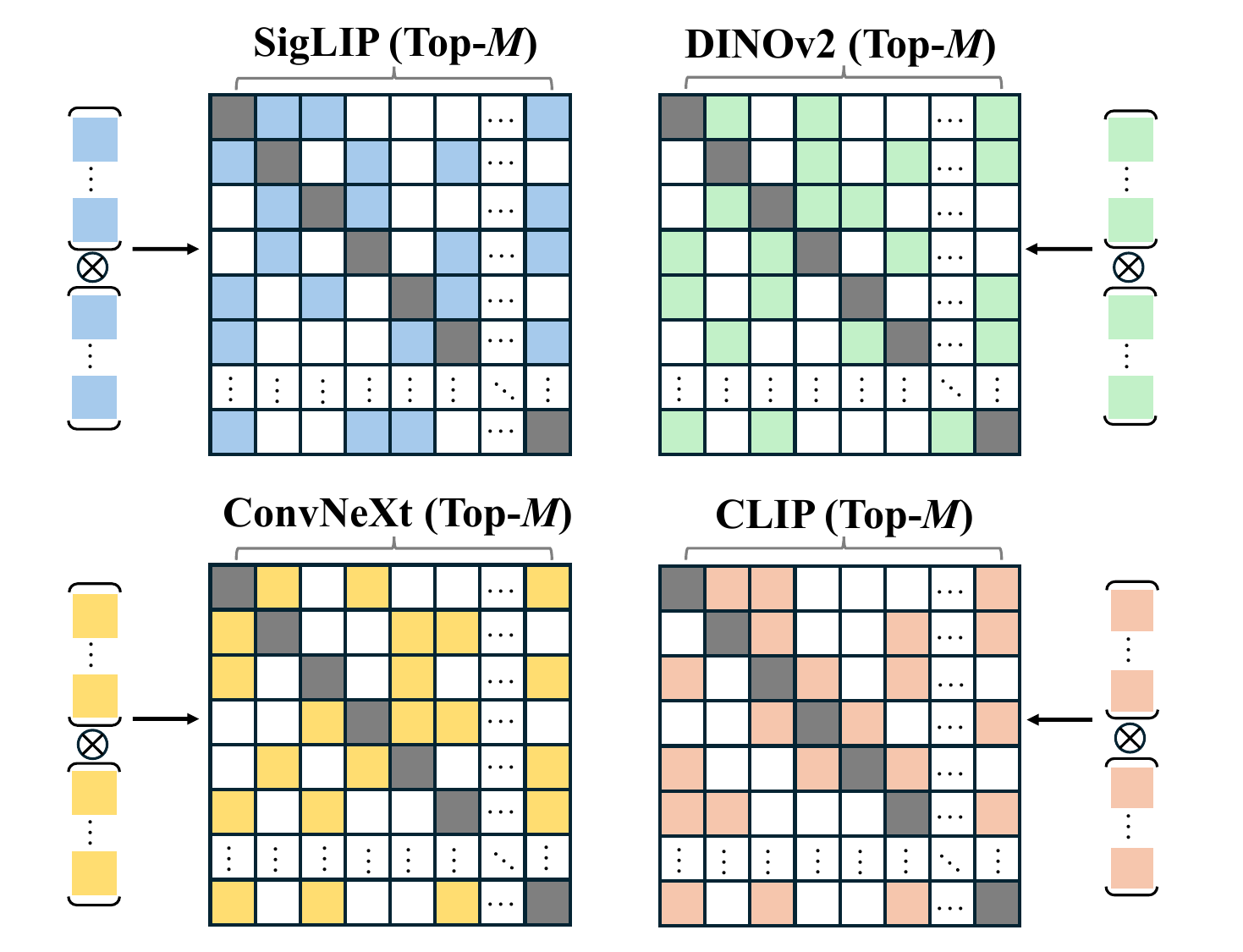}
  \caption{Intra-group Attention. Each model selects and aggregates the Top-$M$ features for its tokens within the same group.}
       \label{fig:mask1}
\end{figure}
\begin{figure}[!htbp]
  \centering
  \includegraphics[width=0.89\columnwidth]{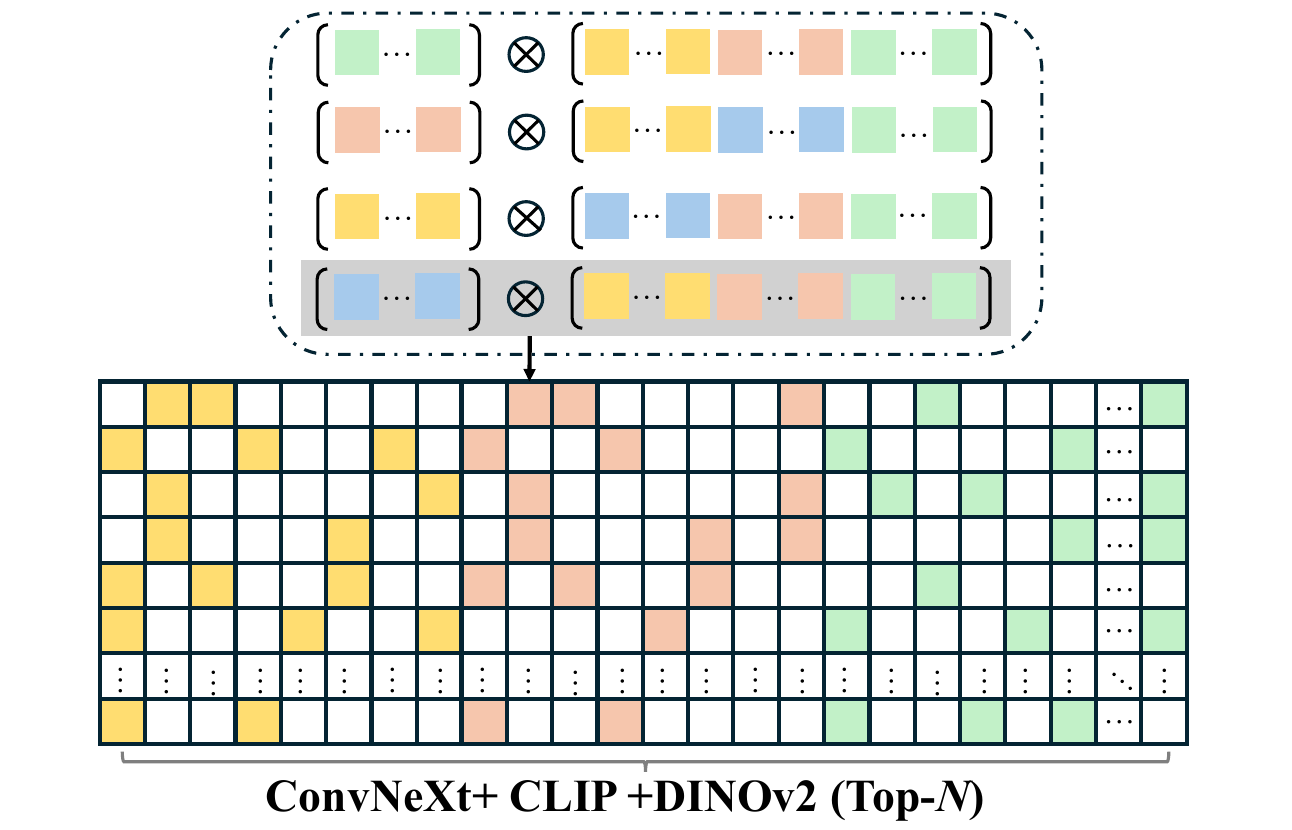}
  \caption{Inter-group Attention. SigLIP selects and aggregates the Top-$N$ features for each token across different models (groups).}
       \label{fig:mask2}
\end{figure}

\subsection{Hierarchical Group Attention}
While sequence append is straightforward to implement, it has limitations, particularly the lack of selective attention to tokens, as all tokens are processed equally, which may introduce redundancy.
To address the above challenge, we propose a hierarchical group attention (HGA) that enables adaptive feature fusion via intra- and inter-group attention, allowing the model to selectively focus on visual tokens.

As illustrated in Figure~\ref{fig:mask1}, output tokens from the four vision encoders (i.e., SigLIP, DINOv2, ConvNeXt and CLIP) are regarded as independent feature groups. The intra-group attention mechanism is used to select the Top-$M$ most salient token features within each group by computing pairwise similarity scores followed by a self-masking operation:
\begin{equation}
I_{\text{intra},s,M}=\text{Top} M(\text{sim}(e_{s,h}, e_{s,h})\odot (1-I)).
\end{equation}

Inter-group attention captures semantic correlations across encoders in high-dimensional space, rather than simple feature redundancy. As illustrated in Figure~\ref{fig:mask2}, it extracts complementary Top-$N$ token information from different encoder features:
\begin{equation}
  I_{\text{inter},s,N}=\text{Top}N(\text{sim}(e_{s,h},[e_{n,h},e_{c,h},e_{d,h}])).
\end{equation}

Hierarchical information from both intra- and inter-group attention is integrated into the aggregated feature representation $X_{in,agg}$ for each token. An adaptive gating mechanism is then applied to balance the contributions of $X_{in,agg}$ and the original features $X_{in}$:
\begin{equation}
\begin{split}
\text{gate}&=\text{Sigmoid}[10*(X_{\text{in,agg}}-X_{\text{in}}-0.2)],\\
X_{\text{out}}&=(1-\text{gate})*X_{\text{in}}+\text{gate}*X_{\text{in,agg}}.
\end{split}
\end{equation}


\subsection{Loss Function}
To maintain a load balance among experts in each MoEC block, we adopt auxiliary losses based on the language modeling cross-entropy loss. This auxiliary loss consists of a load balancing loss~\cite{wang2024auxiliarylossfreeloadbalancingstrategy} and a router z-loss~\cite{zoph2022stmoedesigningstabletransferable}. 
Hence, the total loss is:
\begin{equation}
  \begin{split}
   \mathcal L=\mathcal L_{ce}+\alpha_{b} (\mathcal L_{b,s}+\mathcal L_{b,d}+\mathcal L_{b,n}+\mathcal L_{b,c})\\
   +\alpha_{z} (\mathcal L_{z,s}+\mathcal L_{z,d}+\mathcal L_{z,n}+\mathcal L_{z,c}),
  \end{split}
\end{equation}
where $\mathcal L_{ce}$ represents the autoregressive language modeling loss, 
which computes the cross-entropy of next-token predictions. $\alpha_{b}$ and $\alpha_{z}$ denote coefficients for loading balance loss (i.e., $\mathcal{L}_{b,s}$, $\mathcal{L}_{b,d}$, $\mathcal{L}_{b,n}$, $\mathcal{L}_{b,c}$) and router z-loss (i.e., $\mathcal{L}_{z,s}$, $\mathcal{L}_{z,d}$, $\mathcal{L}_{z,n}$, $\mathcal{L}_{z,c}$), set to 0.1 and 0.01, respectively, across all experiments. These auxiliary losses are applied within the MoECs.
\begin{table*}[!ht]
\centering
\footnotesize
\setlength{\tabcolsep}{2.5pt}
\begin{tabular}{@{}!{\color{white}\vrule width 0pt}l!{\color{black}\vrule width 0.6pt}l!{\color{black}\vrule width 0.6pt}cccccccc!{\color{white}\vrule width 0pt}@{}}
\toprule
Method & LLM & GQA & \makecell{SQA \\IMG} & \makecell{Text \\ VQA} & \makecell{MM\\Vet} & POPE & MME & \makecell{MMB\\EN} & MathVista \\ 
\midrule
Qwen-VL-Chat ~\cite{bai2023qwenvlversatilevisionlanguagemodel}  & Qwen-7B & 57.5 & 68.2 & 61.5 & - & - & 1487.5 & 60.6 & -  \\
InstructBLIP ~\cite{dai2023instructblipgeneralpurposevisionlanguagemodels} & Vicuna-7B &49.2&60.5& 50.1& 26.2& -& -& 36.0& -\\
LLaVA-LLaMA3~\cite{contributorsxtuner}&LLaMA3-8B-IT&62.6& 72.9& 59.0& -& 86.4& 1469& \underline{72.3}& -\\
Mini-Gemini~\cite{li2024minigeminiminingpotentialmultimodality}&Vicuna-7B&-& 65.2& -& 40.8& -& 1523& 69.3& 31.4\\
SPHINX-Intern2~\cite{liu2025sphinxxscalingdataparameters}  & InternLM2-7B& 56.2 & 70.4& 58.1 & 36.5 & 86.9 & 1260.4 & 57.9 & 35.5 \\
LLaVA-NeXT~\cite{liu2024llavanext} &Vicuna-7B&64.2& 70.1& 64.9& 43.9& 86.5& 1519& 67.4& 34.6\\
LLaVA-NeXT~\cite{liu2024llavanext} &Mistral-7B&64.8&72.8&65.7&47.3&86.7&1498&68.7&\textbf{37.7}\\
LLaVA-v1.5~\cite{liu2023visual} &Vicuna-7B&62.0& 66.8& 58.2& 30.5& 85.9& 1510.7& 64.3& -\\
VILA~\cite{lin2024vilapretrainingvisuallanguage} &Vicuna-7B&62.3& 68.2& 64.4& 34.9& 85.5& 1533& 68.9& -\\
CuMo~\cite{li2024cumoscalingmultimodalllm}& Mistral-7B& 64.9&	73.9&	\textbf{67.0}&	\textbf{51.0}&	86.7&	1548.6&	\textbf{73.0}&	35.1\\
\hdashline
InstructBLIP~\cite{dai2023instructblipgeneralpurposevisionlanguagemodels} &Vicuna-13B&49.5&63.1&50.7&25.6&78.9&1212.8&-&-\\
LLaVA-v1.5~\cite{liu2023visual}&Vicuna-13B&63.3&71.6&61.3&35.4&85.9&1531.3&67.7&27.6\\
Mini-Gemini~\cite{li2024minigeminiminingpotentialmultimodality} &Vicuna-13B& -&65.9&-&46.0&-&1565&68.5&\underline{37.0}\\
InternVL-Chat~\cite{chen2024internvl} &Vicuna-13B&\textbf{66.6}&-&61.5&-&87.6&1586.4&-&-\\
LLaMA-VID~\cite{li2024llama} &Vicuna-13B&65.0&70.0&-&-&86.0&1542.3&66.6&-\\
SPHINX-Plus~\cite{liu2025sphinxxscalingdataparameters} &LLaMA2-13B&-&\underline{74.2}&65.7&\underline{47.9}&89.1&1457.7&71.0&36.8\\
\hdashline
MobileVLM~\cite{chu2023mobilevlmfaststrong} &MobileLLaMA-2.7B&59.0& 61.0&47.5& -& 84.9& 1288.9& 59.6& -\\
TinyGPT-V~\cite{yuan2024tinygptvefficientmultimodallarge} &Phi2-2.7B&33.6& -& -& -& -& -& -& -\\
LLaVA-Phi~\cite{zhu2024llava} &Phi2-2.7B&-& 68.4& 48.6& 28.9& 85.0& 1335.1& 59.8& -\\
MoE-LLaVA-2.7B×4-Top2~\cite{lin2024moellavamixtureexpertslarge} &Phi2-2.7B&61.4& 68.5& 51.4& 34.3& 86.3& 1423& 65.2& -\\
\hdashline
\rowcolor{lightblue}
\multicolumn{1}{l!{\color{black}\vrule width 0.8pt}}{MoCHA (SigLIP+DINOv2+ConvNeXt+CLIP)} & \multicolumn{1}{l!{\color{black}\vrule width 0.8pt}}{Phi2-2.7B} & 65.24 & 73.24 & 63.13 & 32.74 & \underline{89.95} & \underline{1701.60} & 69.67 & 32.24$^{\dagger}$ \\
\rowcolor{lightblue}
\multicolumn{1}{l!{\color{black}\vrule width 0.8pt}}{MoCHA (SigLIP+DINOv2+ConvNeXt+CLIP)} & \multicolumn{1}{l!{\color{black}\vrule width 0.8pt}}{Vicuna-7B} & \underline{65.73} & \textbf{74.59} & \underline{65.72} & 36.68 & \textbf{89.98}  & \textbf{1744.09} &71.06 & 35.60$^{\dagger}$ \\
\bottomrule
\end{tabular}
\caption{Comparisons between MoCHA and other VLLMs on competitive benchmarks. These models are grouped by the size of the base LLM. Numbers$^{\dagger}$ are averaged by three inference runs of querying GPT API.}
\label{tab:22}
\end{table*}
\begin{table}[!ht]
\centering
\footnotesize
\setlength{\tabcolsep}{1.5pt}
\begin{tabular}{ll
    >{\centering\arraybackslash}m{1.05cm}
    >{\centering\arraybackslash}m{1.05cm}
    >{\centering\arraybackslash}m{1.3cm}
    c
    }
\toprule
Model & LLM & \makecell{Avg.\\Infer.\\Time} & \makecell{Total\\Params} & \makecell{Trainable\\Params} & GFLOPs \\ 
\midrule
LLaVA-v1.5 & Vicuna-13B & 0.81s & 13.35B & 13.05B & 16894.72 \\
LLaVA-v1.5 & Vicuna-7B&	0.44s&	7.06B&	6.76B&	8877.05 \\
InternVL-Chat&	Vicuna-13B&	0.83s&	18.96B&	13.06B&	23334.93 \\
MoE-LLaVA &	Phi2-2.7B&	\textbf{0.41s}&	5.61B&	5.30B&	\textbf{7338.97}\\
MoCHA&	Phi2-2.7B&	0.57s&	\textbf{4.97B}&	\textbf{4.07B}&	12014.64\\
\bottomrule
\end{tabular}
\caption{Quantitative comparison of MoCHA and other VLLMs.}
\label{tab:quantitive}
\end{table}

\section{Experiments}
\subsection{Data Details}
\subsubsection{Training Data.}The training process utilizes a 558K subset of the LAION-CC-SBU dataset with BLIP-generated captions (as in LLaVA-v1.5), followed by fine-tuning on a 665K mixture of instruction-following data from LLaVA-v1.5.

\subsubsection{Evaluation Benchmarks.}We focus on academic VQA datasets, including GQA~\cite{hudson2019gqa}, Science-QA~\cite{lu2022learn}, and TextVQA~\cite{singh2019towards}, as well as instruction-following VLLM benchmarks, such as POPE~\cite{li2023evaluatingobjecthallucinationlarge}, MME~\cite{fu2024mmecomprehensiveevaluationbenchmark}, MMBench~\cite{liu2024mmbenchmultimodalmodelallaround}, and MM-Vet~\cite{yu2024mmvetevaluatinglargemultimodal}. Furthermore, we evaluate the challenging MathVista~\cite{lu2024mathvistaevaluatingmathematicalreasoning} dataset to assess VLLM visual reasoning capabilities.

\subsection{Implementation Details}
\subsubsection{Two-Stage Training.}To enhance training stability, we adopt a two-stage training strategy. More details and hyperparameters of the training can be found in \textbf{Appendix B}.

\subsubsection{Parameter Settings.}We train for one epoch using 8 NVIDIA L40 (48GB) GPUs. During the pre-training stage, the learning rate is set to 1e-3, with a batch size of $8\times 16$ (1 vision encoder), $8\times 8$ (2 vision encoders) and $8\times 4$ (3 or 4 vision encoders). During visual instruction fine-tuning, the learning rate is reduced to 2e-5, with a batch size of $8\times 8$ (1 vision encoder) and $8\times 4 $ or $2$ (2, 3, or 4 vision encoders). Each MoEC has $E=4$ experts, with $K=2$ activated per forward pass. For adaptive feature fusion through intra- and inter-group attention, we aggregate $M=3$ tokens within each group and $N=7$ tokens across groups.

\subsubsection{Training Settings.} We employ pre-trained OpenAI CLIP ViT-L/14@336, SigLIP ViT-L/16@384, OpenCLIP ConvNeXt-XXL@1024, and DINOv2 ViT-L/14@336 as vision encoders. The vision-language connector is implemented as an MoEC, which consists of a router and expert networks. Each expert consists of two linear layers.
For the language model, we utilize Phi2-2.7B and Vicuna-7B-v1.5. 

\subsubsection{Evaluation Settings.}We follow the LLaVA series settings, using greedy decoding for all benchmarks. Data and questions are converted into visual instructions to prompt VLLMs. For GPT API-based evaluation, we use GPT-3.5-turbo for MathVista.
\subsection{Main Results}
We scale the LLM backbone from 2.7B to 7B parameters for both pre-training and fine-tuning, achieving competitive performance. Comparative analysis is conducted against other instruction-tuned VLLMs, categorized by backbone size (e.g., 2.7B, 7B, 8B and 13B).

\subsubsection{MoCHA Stands Out Among Other VLLMs.} In Table~\ref{tab:22}, we evaluate MoCHA on 8 multimodal benchmarks spanning from general (e.g., GQA, MMVet) to math and reasoning multimodal tasks (e.g., SQA IMG, MathVista). 
Under the same LLM and fine-tuning settings, MoCHA (Phi2-2.7B) consistently outperforms LLaVA-Phi and MoE-LLaVA (Phi2-2.7B) across all benchmarks. Moreover, MoCHA (Phi2-2.7B) achieves overall performance comparable to, and in some cases surpassing, many 7B-based VLLMs as well as larger models. Specifically, MoCHA (Phi2-2.7B) outperforms SPHINX-Plus (LLaMA2-13B) by 0.85\% on POPE and 243.9 points on MME. 
These findings indicate that smaller models can exhibit strong visual perception and excel at fine-grained detail recognition.

As we scale up the LLM backbone, MoCHA (Vicuna-7B) demonstrates stronger visual reasoning capabilities than MoCHA (Phi-2 2.7B). Notably, MoCHA (Vicuna-7B) also surpasses models with even larger LLMs across multiple visual benchmarks. For instance, compared to LLaVA-v1.5 (Vicuna-13B), it achieves improvements of 2.99\%, 4.42\%, and 4.08\% on SQA IMG, TextVQA, and POPE, respectively.
Moreover, in contrast to CuMo—which applies a global mixture of experts (MoE) across the LLM, vision encoder, and connector—MoCHA (Vicuna-7B) outperforms CuMo (Mistral-7B) on several key vision benchmarks.
These results provide strong evidence that, compared to the conventional single-vision-encoder paradigm, MoCHA's dynamic MoE design and customized attention mechanisms offer a more effective approach for multimodal learning (ML).
We present more results regarding other visual capabilities such as multi-object recognition in \textbf{Appendix D}.

\subsubsection{MoCHA Helps Reduce Visual Hallucination.} We follow the evaluation protocol of POPE, a popular benchmarking for evaluating visual object hallucination. 
MoCHA shows a marked improvement in mitigating hallucination issues in Table~\ref{tab:22}, outperforming models with larger parameter sizes. It produces object descriptions that are highly consistent with the input images and exhibits a rich potential for hallucination inhibition. Specifically, we observe that MoCHA surpasses LLaVA in popular sampling, adversarial sampling, and random sampling, despite having fewer parameters. Additionally, the yes ratio of MoCHA remains relatively balanced, which indicates that our framework is capable of providing accurate feedback based on the given questions.

\subsubsection{The MoCHA Design Optimizes Model Efficiency.} To validate the efficiency use of our MoECs, we further present training GFLOPs, parameters, and average inference time of models in Table~\ref{tab:quantitive}. 
We make sure that all numbers are obtained under identical configurations to ensure the validity of the comparison. 
From the table, MoCHA achieves an inference speed of 0.57s with only 4.97B parameters and 12014.64 GFLOPs. This not only significantly reduces both the parameter size and computational cost, but also outperforms LLaVA-v1.5 (Vicuna-13B) and InternVL-Chat (Vicuna-13B) in inference efficiency. Furthermore, it approaches the inference speed of MoE-LLaVA, despite using even fewer parameters.
These notable advantages in both computational and parameter efficiency highlight the strong potential of scaling up MoCHA as a compelling alternative to existing vision encoder paradigms in ML.
\begin{figure}[!htbp]
\centering
\includegraphics[width=0.99\linewidth]{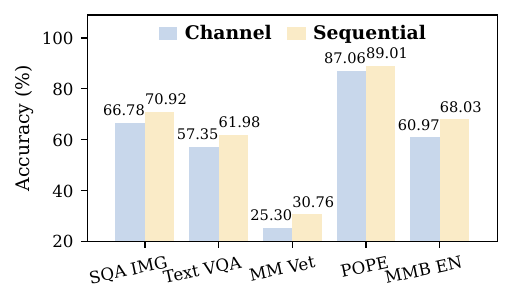} 
\caption{Ablation of Channel vs. Sequential Concatenation for MoECs with four encoders.}
\label{concat}
\end{figure}
\subsection{Ablation Studies}
\subsubsection{Effect of Channel v.s. Sequential Concatenation.} We conduct ablation experiments on the heterogeneous features dynamically selected by the MoECs module with sequential concatenation and channel concatenation to evaluate the differences between the two strategies.
As illustrated in Figure~\ref{concat}, sequence concatenation consistently outperforms channel concatenation across all vision tasks under our MoEC settings. Furthermore, channel concatenation aggregates features from multiple encoders within each visual token; if one encoder (e.g., CLIP) has dominant feature magnitudes, it may overshadow fine-grained information from other encoders, leading the model to focus disproportionately on its features.

\begin{table}[t]
\centering
\footnotesize
\setlength{\tabcolsep}{2.5pt}
\begin{tabular}{l!{\color{black}\vrule width 0.6pt}cccc}
\toprule
  Method (MLP)& GQA &\makecell{SQA \\IMG}&POPE & \makecell{MMB\\EN} \\ 
\midrule
SigLIP& 59.00&	64.35&	84.64&	59.88 \\
        SigLIP+ConvNeXt &60.78&	67.18&	86.67&	64.95      \\
        SigLIP+CLIP+ConvNeXt&57.21&	65.39&	85.72&	58.42 \\
        SigLIP+DINOv2+ConvNeXt+CLIP&\textbf{61.86}&	\textbf{69.01}&	\textbf{88.47}&\textbf{66.51}\\       
\bottomrule
\end{tabular}
\caption{Ablation study only on different vision encoder combinations.}
\label{tab:group}
\end{table}
\begin{table}[htbp]
\centering
\footnotesize
\setlength{\tabcolsep}{2.5pt}
\begin{tabular}{l!{\color{black}\vrule width 0.6pt}cccc}
\toprule
  Method (MoEC) & GQA &\makecell{SQA \\IMG}&POPE & \makecell{MMB\\EN} \\ 
\midrule
SigLIP& 56.77&	64.60&	82.85&	57.65 \\
        SigLIP+ConvNeXt &57.77&	67.92&	86.72&63.49      \\
        SigLIP+CLIP+ConvNeXt&59.10&	66.58&	87.06&	60.83 \\
        SigLIP+DINOv2+ConvNeXt+CLIP&\textbf{63.35}&	\textbf{70.92}&	\textbf{89.01}&\textbf{68.03}\\       
\bottomrule
\end{tabular}
\caption{Ablation study of different vision encoder combinations under MoEC settings.}
\label{tab:moe}
\end{table}
\subsubsection{Effect of Multiple Vision Encoders.} 
To investigate the use of different combinations of vision encoders under the MLP connector,
ablation studies were conducted by individually extracting features from each vision encoder. 
As shown in Table~\ref{tab:group}, our study indicates that only using the SigLIP exhibits inferior performance across various benchmarks. 
The inclusion of ConvNeXt improves performance on benchmarks such as GQA, SQA IMG, and POPE. However, the subsequent addition of CLIP leads to a performance decline. 
This phenomenon can be attributed to the substantial feature similarity and redundancy between SigLIP and CLIP, since both encoders are trained with large-scale image-text contrastive objectives. The configuration that integrates SigLIP, DINOv2, ConvNeXt, and CLIP achieves the best overall performance. In particular, DINOv2 provides a balanced contribution by capturing both local details and global semantic information.

\subsubsection{Upgrade MLP connector to MoEC.} 
To evaluate MoEC’s capability in efficiently integrating heterogeneous and multi-dimensional features,
we initiate this ablation study by replacing each MLP connector with an upcycled MoEC, as shown in Table~\ref{tab:moe}. 
The results in Table~\ref{tab:group} indicate that with the plain MLP as the connector, the model performance with a single SigLIP remains weak. While the addition of ConvNeXt further improves performance, the full integration of CLIP yields the largest gain among all settings. 
On the other hand, in Table~\ref{tab:moe}, the combination of SigLIP, DINOv2, ConvNeXt, and CLIP under our MoEC results in notable performance improvements of 1.49\%, 1.91\%, 0.54\%, and 1.52\% on the GQA, SQA IMG, POPE, and MMBench benchmarks, respectively, compared to results using vanilla MLP in Table~\ref{tab:group}. 
These findings show that MoECs effectively handle diverse visual features, including object recognition, scene understanding, attribute identification, and spatial reasoning. Even as the number of vision encoders grows, MoECs incur minimal parameter overhead while keeping inference time and computational cost on par with a standard MLP. Further quantitative results are provided in \textbf{Appendix C}.
\begin{table}[htbp]
\centering
\footnotesize
\setlength{\tabcolsep}{2.5pt}
\begin{tabular}{l!{\color{black}\vrule width 0.6pt}cccc}
\toprule
Method (MoEC + HGA) & GQA & \makecell{SQA\\IMG} & POPE & \makecell{MMB\\EN} \\ 
\midrule
SigLIP & 56.11 & 62.76 & 82.27 & 57.90 \\
SigLIP+ConvNeXt & 57.02 & 63.97 & 87.84 & 61.06 \\
SigLIP+CLIP+ConvNeXt & 60.34 & 68.73 & 88.07 & 62.11 \\
\rowcolor{lightblue} \multicolumn{1}{l!{\color{black}\vrule width 0.8pt}} {SigLIP+DINOv2+ConvNeXt+CLIP }& \textbf{65.24} & \textbf{73.24} & \textbf{89.95} & \textbf{69.67} \\
\bottomrule
\end{tabular}
\caption{Ablation Study of MoCHA with different vision encoder combinations. Settings for results in Table~\ref{tab:22} are highlighted in \colorbox{lightblue}{blue}.}
\label{tab:graph}
\end{table}
\begin{table}[htbp]
\centering
\footnotesize
\setlength{\tabcolsep}{3pt}
\begin{tabular}{l!{\color{black}\vrule width 0.6pt}cccc}
\toprule
  Top-$K$ & GQA &\makecell{SQA \\IMG}&POPE & \makecell{MMB\\EN} \\ 
\midrule
1 &59.86 &68.13 &85.93 & 65.59\\
 \rowcolor{lightblue} \multicolumn{1}{l!{\color{black}\vrule width 0.8pt}} 2 & {\textbf{65.24}}&	\textbf{73.24}&	\textbf{89.95}&	\textbf{69.67} \\   
 3& 62.90&	70.18&	88.21&	67.35\\
 4& 58.32&	67.64&	86.02&	64.97\\
\bottomrule
\end{tabular}
\caption{Ablation study on the value of Top-$K$ in MoCHA.}
\label{tab:expertnum}
\end{table}
\subsubsection{Effect of Hierarchical Group Attention (HGA).} We further ablate our hierarchical group attention (HGA) to evaluate its effectiveness in feature fusion.
As shown in Table~\ref{tab:graph}, continual performance gains are observed as ConvNeXt, CLIP, and DINOv2 are introduced sequentially. This confirms that the proposed intra- and inter-group adaptive attention fusion in HGA enhances the synergy and complementarity among multiple vision encoders.
Compared to previous results, the performance of the single SigLIP encoder slightly drops relative to both MoEC (Table~\ref{tab:moe}) and the standard MLP baseline (Table~\ref{tab:group}), suggesting that intra-group feature aggregation within a single encoder may introduce redundancy.
Notably, incorporating ConvNeXt leads to a 5.57\% improvement on the POPE benchmark, indicating that HGA effectively leverages ConvNeXt’s fine-grained local features.
The full combination of SigLIP, DINOv2, ConvNeXt, and CLIP yields the best overall performance, further demonstrating the robustness of MoCHA.

\subsubsection{Effect of the Number of Activated Experts.}To evaluate the effect of the number of activated experts, we compare different Top-$K$ strategies. As shown in Table~\ref{tab:expertnum}, $K=2$ achieves the best performance, while larger $K$ values provides limited improvement, increases computational and memory costs, lowers parameter utilization, and may cause expert imbalance. Hardware constraints further restrict $K=4$ to a batch size of 1, which may degrade performance. Accordingly, we set $K=2$ to fully exploit the advantages of the MoEC architecture.

\section{Conclusion}
In this work, we introduce MoCHA, a novel framework for efficiently training vision-language models with advanced visual reasoning capabilities. It addresses two key challenges in visual detail extraction and in heterogeneous feature fusion. 
The MoECs module enables dynamic expert selection, effectively integrating diverse visual signals within a unified vision-language framework. Customized HGA further enhances the complementarity of heterogeneous features.
Our framework can be readily adapted to incorporate additional vision backbones and larger language models.
Although MoCHA demonstrates remarkable performance, MoEC may suffer from knowledge entanglement and redundancy, which could hinder expert specialization. Future work may explore fine-grained expert partitioning and shared expert isolation to further improve knowledge allocation within the model. 
We hope this work provides a foundation and new insights for vision encoder design in VLLMs.

\bibliography{aaai2026}

\clearpage
\twocolumn
\appendix
\section{Appendix}
\subsection{A Analysis of Vision Encoders}
\label{vision}
A single vision encoder is unlikely to achieve the best performance across all benchmark tasks. As shown in Table~\ref{tab:single}, DINO performs best on the GQA benchmark, CLIP achieves the highest score on MMBench, while SigLIP demonstrates advanced performance on MME and MM-Vet tasks. These results indicate that different vision encoders exhibit complementary strengths in visual reasoning tasks.

CLIP achieves vision-language alignment through language supervision, SigLIP employs a binary classification objective to improve training efficiency and zero-shot performance, ConvNeXt leverages a convolutional architecture for high-resolution local feature extraction, and DINOv2 captures pixel-level geometric structure via self-supervised learning. These backbones exhibit fundamental differences in
architecture (CNN vs. ViT), training paradigms (supervised vs. self-supervised), and information granularity (global vs. local), leading
to naturally complementary feature representations that enhance
both diversity and robustness.

\begin{table}[!h]
\centering
\footnotesize
\setlength{\tabcolsep}{2pt}
\begin{tabular}{l!{\color{black}\vrule width 0.6pt}l!{\color{black}\vrule width 0.6pt}cccc}
\toprule
 Vision Model & LLM& GQA &\makecell{MMB \\EN}&MME & \makecell{MM\\Vet} \\ 
\midrule
DINO&	Phi2-2.7B&	\textbf{61.00}&	57.82&	1555.61&	22.3\\
ConvNeXt&	Phi2-2.7B&	58.51&	64.00&	1604.99	&24.5\\
CLIP&	Phi2-2.7B&	60.12&	\textbf{66.92}&	1600.19	&25.4\\
SigLIP&	Phi2-2.7B&	59.00&	59.88&	\textbf{1622.61}	&\textbf{25.7}\\
\bottomrule
\end{tabular}
\caption{Evaluation performance of individual vision encoders.}
\label{tab:single}
\end{table}

\subsection{B Train Details}   
\label{trian}
\begin{figure}[ht]
  \centering
  \includegraphics[width=\columnwidth]{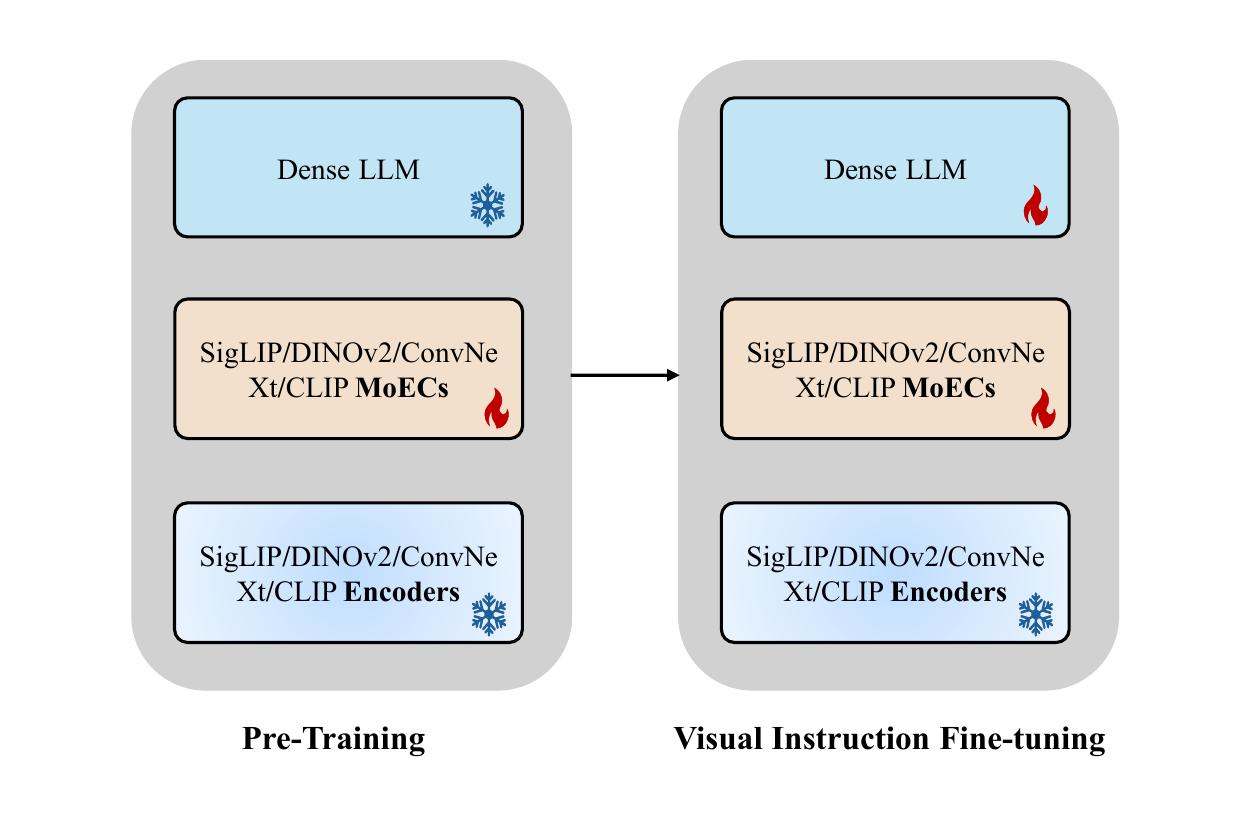}
  \caption{Training stages of MoCHA.}
       \label{fig:tain}
\end{figure}
\begin{table}[!ht]
\centering
  \footnotesize
\setlength{\tabcolsep}{1mm}
\begin{tabular}{l!{\color{black}\vrule width 0.6pt}cc}
\toprule
Config & Stage I & Stage II \\
\midrule
Experts              & --       & --    \\
Top-k                & --       & 2     \\ \midrule
Image encoder        & \multicolumn{2}{c}{SigLIP, DINOv2, ConvNeXt, CLIP} \\
Feature select layer & \multicolumn{2}{c}{-2} \\
Image projector      & \multicolumn{2}{c}{MoECs (Multiple MoEC units)} \\
Image resolution &  \multicolumn{2}{c}{e.g., 384, 1024, 336}\\
Epoch                & \multicolumn{2}{c}{1} \\
Learning rate        &  1e-3     &  2e-5  \\
Learning rate schdule& \multicolumn{2}{c}{Cosine} \\
Weight decay         & \multicolumn{2}{c}{0.0} \\
Text max length      & \multicolumn{2}{c}{2560} \\
Batch size per GPU   &   4      &     2   \\
GPU                  & \multicolumn{2}{c}{8 $\times$ L40-48GB} \\
Precision            & \multicolumn{2}{c}{Bf16} \\
\bottomrule
\end{tabular}
\caption{Training hyperparameters.}
\label{training_hypers}
\end{table}

As illustrated in Figure~\ref{fig:tain}, we now describe the corresponding MoCHA training strategy, which consists of three stages:
\begin{itemize}
\item \textbf{Stage 1: Pre-training for Feature Alignment. }We freeze the vision encoders and the large language model, pre-training only the SigLIP MoEC, DINOv2 MoEC, ConvNeXt MoEC, and CLIP MoEC. 
\item \textbf{Stage 2: Visual Instruction Fine-tuning. }The weights of all four vision encoders remain frozen, while the MoECs and LLM pre-trained weights are further updated.
\end{itemize}
\begin{table*}[!ht]
\centering
\footnotesize
\setlength{\tabcolsep}{2.5pt}
\begin{tabular}{@{}lccccc@{}}
\toprule
Vision Model & Total Tokens & \makecell{Avg.Infer.Time\\(MLP vs MoEC)} & \makecell{Total Params \\ (MLP vs MoEC)} & \makecell{Trainable Params \\ (MLP vs MoEC)} & \makecell{GFLOPs \\ (MLP vs MoEC)} \\
\midrule
SigLIP & 440 & 0.34s vs 0.34s & 3.38B vs 3.41B & 2.78B vs 2.81B & 3000.09 vs 3008.17 \\
SigLIP+ConvNeXt & 540 & 0.41s vs 0.31s & 4.23B vs 4.29B & 3.64B vs 3.69B & 4244.34 vs 4254.26 \\
SigLIP+CLIP+ConvNeXt & 1116 & 0.45s vs 0.39s & 4.24B vs 4.33B & 3.65B vs 3.73B & 8021.32 vs 8048.25 \\
SigLIP+DINOv2+ConvNeXt+CLIP & 1692 & 0.58s vs 0.57s & 4.86B vs 4.97B & 3.96B vs 4.07B & 11983.58 vs 12014.64 \\
\bottomrule
\end{tabular}
\caption{Ablation study on inference efficiency and model complexity across single and multi-encoder variants.}
\label{tab:ablation_efficiency}
\end{table*}

As shown in Table~\ref{training_hypers}, we present the training hyperparameters for all models, which are applicable to Phi and Vicuna. For the training process in all stages, we consistently train for 1 epoch, as we find that the models overfit when training for 2 epochs. We enable the gradient checkpoint mode for all training stage.

\subsection{C Quantitative Analysis of Different Vision Encoder Combinations}
\label{quantitative}
We conduct detailed statistics on inference time, total parameters, trainable parameters, and GFLOPs when using either MLP or MoEC modules across single and multi-encoder variants, all evaluated on the same dataset.    
As shown in Table~\ref{tab:ablation_efficiency}, the results demonstrate that MoEC significantly improves inference efficiency in multi-encoder settings (e.g., reducing inference time from 0.41s to 0.31s with the SigLIP+ConvNeXt combination), while introducing only minimal increases in total and trainable parameters (e.g., 4.97B/4.07B vs. 4.86B/3.96B in the SigLIP+DINOv2+ConvNeXt+CLIP configuration). MoECs can seamlessly support the integration of multiple vision encoders, owing to their sparse gating and expert allocation mechanisms. As a result, features from different vision encoders can be flexibly routed to the most appropriate experts for effective fusion. Notably, the hierarchical group attention module achieves adaptive feature fusion without adding extra trainable parameters. We also evaluate other VLLMs and our MoCHA on the same dataset, with identical batch size and hardware.

\subsection{D More Results}
As illustrated in Figure~\ref{fig:example1}, 
both MoCHA and LLaVA provide accurate answers when identifying that the person is riding an elephant rather than a horse or determining the position of the curtain. Both MoCHA and LLaVA demonstrate strong capabilities across diverse reasoning tasks, such as fundamental visual object recognition and simple spatial relationship inference.

As illustrated in Figure~\ref{fig:example2}, MoCHA (Vicuna-7B) demonstrates advanced reasoning and comprehension abilities in multi-object recognition and fine-grained visual reasoning tasks. For example, it accurately identifies the position of unsweetened milk and the number on the motorcycle ridden by the motorcyclist in a yellow suit. In contrast, LLaVA (Vicuna-7B) and MoCHA (Phi2-2.7B) exhibit certain limitations in understanding complex scenes. We highlight that this improvement is not attributable to increased language model size, but rather to the effective enhancement of visual information introduced by the MoCHA framework.

\begin{table}[htbp]
\centering
\footnotesize
\begin{tabularx}{\columnwidth}{l X}
\toprule
\multicolumn{2}{l}{\textbf{Visual input example, Question and Image:}} \\
\midrule
\multicolumn{2}{c}{\includegraphics[width=0.8\linewidth,height=4cm,keepaspectratio]{v1_130.jpg}} \\
\addlinespace[2pt]
\textbf{User} & Does the man ride a horse?\\
\addlinespace[2pt]
\textbf{LLaVA (Vicuna-7B)} & No, the man does not ride a horse. He is riding on the back of an elephant. \bmark \\
\addlinespace[2pt]
\textbf{MoCHA (Phi2-2.7B)} & No, the man does not ride a horse. He is riding an elephant. \bmark \\
\addlinespace[2pt]
\textbf{MoCHA (Vicuna-7B)} & No, the man does not ride a horse. He is riding an elephant. \bmark\\
\midrule
\multicolumn{2}{c}{\includegraphics[width=0.8\linewidth,height=4cm,keepaspectratio]{v1_122.jpg}} \\
\addlinespace[2pt]
\textbf{User} & Is the curtain on the right side or on the left of the picture? \\
\addlinespace[2pt]
\textbf{LLaVA (Vicuna-7B)} & The curtain is on the left side of the picture. \bmark \\
\addlinespace[2pt]
\textbf{MoCHA (Phi2-2.7B)} & The curtain is on the left side of the picture. \bmark \\
\addlinespace[2pt]
\textbf{MoCHA (Vicuna-7B)} & The curtain is on the left side of the picture. \bmark\\
\bottomrule
\end{tabularx}
\caption{Example comparing LLaVA and MoCHA's visual reasoning capabilities in understanding fundamental secene.}
\label{fig:example1}
\end{table}

\begin{table}[htbp]
\centering
\footnotesize
\begin{tabularx}{\columnwidth}{l X}
\toprule
\multicolumn{2}{l}{\textbf{Visual input example, Tricky Question and Image:}} \\
\midrule
\multicolumn{2}{c}{\includegraphics[width=0.8\linewidth,height=4cm,keepaspectratio]{v1_67.png}} \\
\addlinespace[2pt]
\textbf{User} & Which one is unsweetened?\\
\addlinespace[2pt]
\textbf{LLaVA (Vicuna-7B)} & The unsweetened milk is on the left side of the image. \bmark\\
\addlinespace[2pt]
\textbf{MoCHA (Phi2-2.7B)} & \textcolor{red}{The unsweetened milk is the one that is not sweetened.} \xmark \\
\addlinespace[2pt]
\textbf{MoCHA (Vicuna-7B)} & The unsweetened milk is the one on the left. \bmark\\
\midrule
\multicolumn{2}{c}{\includegraphics[width=0.8\linewidth,height=4cm,keepaspectratio]{v1_28.jpg}} \\
\addlinespace[2pt]
\textbf{User} & What is the number displayed on the motorcycle ridden by the motorcyclist wearing a yellow suit?\\
\addlinespace[2pt]
\textbf{LLaVA (Vicuna-7B)} & The number displayed on the motorcycle ridden by the motorcyclist wearing a yellow suit is 19. \xmark\\
\addlinespace[2pt]
\textbf{MoCHA (Phi2-2.7B)} & \textcolor{red}{The number displayed on the motorcycle on the right is 21. } \xmark\\
\addlinespace[2pt]
\textbf{MoCHA (Vicuna-7B)} &The motorcyclist wearing a yellow suit is riding a motorcycle with the number 16 on it.\bmark\\
\bottomrule
\end{tabularx}
\caption{Example comparing LLaVA and MoCHA's visual reasoning capabilities in understanding multi-object recognition. }
\label{fig:example2}
\end{table}

\end{document}